\documentclass[runningheads]{llncs}
\usepackage[T1]{fontenc}
\usepackage{graphicx}
\usepackage{amsmath}
\usepackage{amssymb}
\usepackage{multirow}
\usepackage{booktabs}

\begin{document}
\title{Semi-Disentangled Spatiotemporal Implicit Neural Representations of Longitudinal Neuroimaging Data for Trajectory Classification}
\titlerunning{Implicit Neural Representations for Trajectory Classification}

\author{Agampreet Aulakh\inst{1} \and
Nils D. Forkert\inst{2} \and
Matthias Wilms\inst{3}}

\authorrunning{Aulakh et al.}
\institute{Biomedical Engineering Graduate Program, University of Calgary, Calgary, Canada \and
Department of Radiology, University of Calgary, Calgary, Canada \and
Department of Radiology, University of Michigan, Ann Arbor, United States
}

\maketitle              
\begin{abstract}
The human brain undergoes dynamic, potentially pathology-driven, structural changes throughout a lifespan. Longitudinal Magnetic Resonance Imaging (MRI) and other neuroimaging data are valuable for characterizing trajectories of change associated with typical and atypical aging. However, the analysis of such data is highly challenging given their discrete nature with different spatial and temporal image sampling patterns within individuals and across populations. This leads to computational problems for most traditional deep learning methods that cannot represent the underlying continuous biological process. To address these limitations, we present a new, fully data-driven method for representing aging trajectories across the entire brain by modelling subject-specific longitudinal T1-weighted MRI data as continuous functions using Implicit Neural Representations (INRs). Therefore, we introduce a novel INR architecture capable of partially disentangling spatial and temporal trajectory parameters and design an efficient framework that directly operates on the INRs' parameter space to classify brain aging trajectories. To evaluate our method in a controlled data environment, we develop a biologically grounded trajectory simulation and generate T1-weighted 3D MRI data for 450 healthy and dementia-like subjects at regularly and irregularly sampled timepoints. In the more realistic irregular sampling experiment, our INR-based method achieves 81.3\% accuracy for the brain aging trajectory classification task, outperforming a standard deep learning baseline model (73.7\%).

\end{abstract}
\section{Introduction}
Subject-specific longitudinal neuroimaging data, such as repeated T1-weighted Magnetic Resonance Imaging (MRI), are vital for data-driven studies of healthy brain aging and pathological deviations associated with neurodegenerative diseases. These data enable the characterization of subject- and population-specific trajectories of typical and atypical brain changes over time. For example, in the case of Alzheimer's Disease (AD), longitudinal data have been instrumental in identifying the morphological signatures of different disease subtypes \cite{poulakis2022multi} and can also be used to detect conversion from normal aging to dementia early on \cite{cui2019rnn}.

Despite their value and increasing availability through long-term studies, such as the Alzheimer's Disease Neuroimaging Initiative (ADNI) cohorts \cite{mueller2005alzheimer}, acquiring longitudinal images in a fully uniform manner across larger cohorts is extremely difficult. In practice, the number of images acquired for each participant usually differs, the data are typically sampled at inconsistent timepoints, and if acquisition is conducted over decades, it is reasonable to expect considerable differences in spatial resolution and image quality over time \cite{ulrich2025back}. All of these aspects make the efficient and meaningful processing of information available at the discretized image level highly challenging.

To circumvent some of these issues, a substantial body of research has been dedicated to extracting scalar image-derived biomarkers, such as cortical thickness or ventricular volume, to analyze brain aging trajectories through mixed-effects models (\textit{e.g.,} \cite{mofrad2021cognitive}). While such methods produce clinically relevant results and are computationally efficient, they lack the flexibility needed to capture the full range of possible subject-specific trajectories and heavily rely on arbitrary design decisions, such as the parcellation scheme used for biomarker extraction. Moreover, these methods only give rise to coarse, biomarker-level trajectories and do not capture important interactions between image features.

Beyond traditional biomarker analyses, several deep learning methods have been developed to model brain aging trajectories directly using imaging data. For instance, it has been shown that trajectories can be modelled as embeddings between image pairs within a latent space that is organized by subjects with similar dynamics \cite{ouyang2021self,ouyang2022self}. Other methods \cite{cui2019rnn,ouyang2020longitudinal} first use neural networks to extract features from each image within a longitudinal dataset, and then classify each timepoint as healthy or dementia-like. Critically, these approaches are not able to directly parameterize the full longitudinal aging trajectory \cite{ouyang2021self,ouyang2022self}, or rely on intermediate, imaging feature-based representations for analyses \cite{cui2019rnn,ouyang2020longitudinal}.

We aim to address the challenges associated with processing variable, subject-specific, longitudinal neuroimaging data and the shortcomings of current deep learning-based analysis methods in a unified way to make the best use of the full discrete image information available. Specifically, we propose to directly parameterize individual brain aging trajectories by modelling longitudinal imaging data in a space- and time-continuous manner via Implicit Neural Representations (INRs). These INRs are simple neural networks that learn to approximate a continuous function mapping from the domain of coordinates to their associated signal value. Such data representations have been previously adapted to and used for the classification of 3D object images/point clouds \cite{dupont2022data,luigi2023deep}, high-resolution natural images \cite{gielisse2025end,bauer2023spatialfunctascalingfuncta}, and, more recently in a medical context, low-resolution 2D scans \cite{friedrich2025medfuncta}. Lately, INRs have also been popularized for medical image registration purposes \cite{wolterink2022implicit,shuaibu2025capturing}. Applied to longitudinal neuroimaging, the space- and time-continuous nature of INRs overcomes the problems associated with different temporal (and spatial) sampling patterns and enables a fully data-driven, learning-based representation of longitudinal data. INRs can be flexibly adapted to subject-specific data as the parameter count of these networks is independent of the number of subjects studied and amount of data available per subject. Furthermore, we show for the first time that their parameters can be used for the classification of brain aging trajectories (normal aging \textit{vs.} AD-like aging).

Thus, the main contributions of this paper are threefold: First, we present a new INR architecture with semi-disentangled spatial and temporal parameters to model subject-level aging trajectories across the entire brain from longitudinal T1-weighted MRI data. We show that these INRs can be constructed using data with irregular temporal sampling within subjects and across a cohort. Second, we propose an efficient method for directly classifying the trajectories encoded by these representations. Third, to perform a thorough proof-of-concept evaluation of our methods in a fully controlled manner, we introduce a synthetic brain aging trajectory simulation framework capable of generating longitudinal 3D imaging data of realistic healthy and AD-like aging patterns.

\begin{figure}[]
\includegraphics[width=\textwidth]{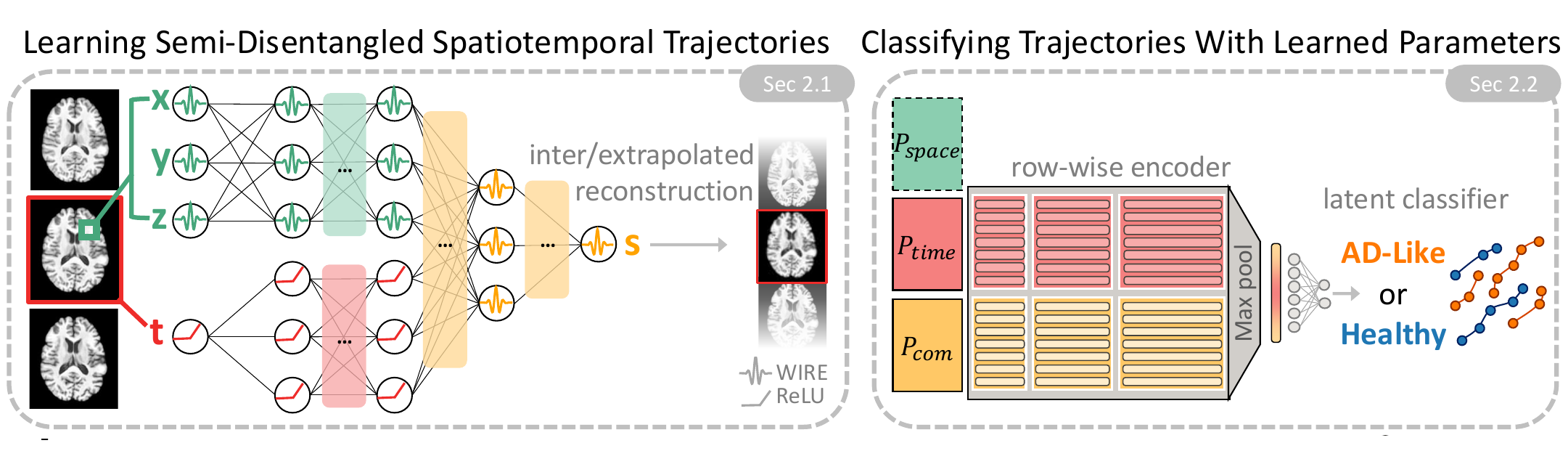}
\caption{Overview of proposed methods. Left: Subject-specific trajectory modelling with a semi-disentangled spatiotemporal INR. Right: Organization of INR parameters in space-only, time-only, and combined streams, their row-wise embedding in latent space, and subsequent trajectory classification.} \label{fig:wf}
\end{figure}

\section{Materials and Methods}
We assume the biological process of brain aging follows a continuous, subject-specific trajectory \(A(\mathbf{c},t)\) and consider individual T1-weighted MRI scans (longitudinal data) as discrete samples of the structural changes over time $t$ at each spatial coordinate $\mathbf{c}$. Our goal is then twofold: (1) To continuously approximate the underlying trajectory in space and time through an INR trained on discrete image samples. (2) To build a computationally efficient trajectory classifier that operates on the continuous INRs. 

Formally, we consider data from $M$ subjects such that each subject \(m \in \{1,\dots,M\}\) has a temporal sequence of \(N_m\) scans denoted as \(\mathcal{D}=\{(\mathbf{I}_i,t_i)\}_{i=1}^{N_m}\). Here, \(\mathbf{I}_i \in \mathbb{R}^{X\times Y \times Z}\) is the 3D image tensor acquired at a discrete sampling time \(t_i \in \mathcal{T} \subset \mathbb{R}\). Each image corresponds to a discrete sampling pattern of spatial coordinates \(\mathbf{c} =(x,y,z)^T \in \mathcal{C} \subset \mathbb{R}^3\), with voxels representing T1-weighted MRI signal intensity values \(s \in \mathcal{S} \subset \mathbb{R}\). While this data $\mathcal{D}$ is inherently discrete, we derive a continuous approximation of the brain aging trajectory \(f_{\theta}(\mathbf{c},t) \approx A(\mathbf{c},t)\) through an INR $f_{\theta}(\cdot)$ with parameters \(\theta\) that fully parameterize the underlying continuous process (see Sec. \ref{sec:inr}). Assuming a fixed INR architecture across all $M$ subjects, we then propose a trajectory classifier that uses the subject-specific parameters \(\theta_{m}\) as input (see Sec. \ref{sec:class}). The framework is illustrated in Fig. \ref{fig:wf}.

\subsection{Semi-Disentangled Spatiotemporal Implicit Neural Representations}\label{sec:inr}
Following common practice \cite{saragadam2023wire}, we use a multilayer perceptron (MLP) to instantiate an INR. Specifically, we construct a subject-specific INR \(f_{\theta}:\mathcal{C} \times \mathcal{T} \rightarrow \mathcal{S}\) mapping the input image domain tuples \((\mathbf{c},t)\) to image signal values \(s \in \mathcal{S}\). For an \(L\)-layer INR with hidden size \(H\), we use residual connections such that the output of each layer $l$ is defined as:
\begin{equation}
    \textbf{y}_l = \sigma \left(\mathbf{W}_l\textbf{y}_{l-1} + \mathbf{b}_l \right)\,+\,\textbf{y}_{l-1},
\end{equation}
where \(\sigma\) is an activation function, and \(\mathbf{W}_l \in \mathbb{R}^{H\times H}\) and \(\mathbf{b}_l \in \mathbb{R}^{H\times 1}\) are the weights and biases. These parameters are optimized for a subject $m$ by minimizing an MSE loss \(\mathcal{L}=\frac{1}{J\cdot N_m}\sum_{i=1}^{N_m}\sum_{j=1}^{J}(f_{\theta}(\mathbf{c}_j,t_i)-\mathbf{I}_i[\mathbf{c}_j])^2\) over a randomly selected subset of \(J\) voxels for each \(\mathbf{I}_i\) in \(\mathcal{D}\).

While fitting an INR to a longitudinal dataset \(\mathcal{D}\) is generally challenging due to its usually high spatial and low temporal resolution, we make two key observations to simplify this optimization task: (1) Neuroimaging data contains high spatial frequencies as regions of the brain have unique structures (\textit{e.g.,} cortical folding). (2) The frequencies of the temporal domain remain low as one can expect gradual, aging or pathology-driven structural change over time.

We, therefore, propose to disentangle the spatial and temporal processing early on by separating layers receiving $\mathbf{c}$ as input from those receiving $t$, and to only combine spatial and temporal information at the end (see Fig. \ref{fig:wf}). This results in a novel, semi-disentangled INR architecture where each stream can be independently biased to capture a different range of frequencies using activation functions $\sigma$. Namely, we use wavelet activations (\textit{i.e.}, WIRE \cite{saragadam2023wire}) for the spatial and combined streams as they are the state-of-the-art for capturing high-frequency signals, and use ReLU activations for the temporal stream. The spectral biases of both activations have been studied in the past, with multiple works providing results of ablation experiments that complement our design \cite{essakine2024we,rahaman2019spectral,saragadam2023wire}. With this architecture, we partially decouple space-only, time-only, and space-time combined parameters, which enables us to selectively ignore or emphasize each stream in downstream analysis (\textit{e.g.}, trajectory classification). This is advantageous for limiting morphology bias (\textit{e.g.,} exploiting shortcuts in space-only parameters) when studying biological processes that are largely time-dependent, such as \(A(\mathbf{c},t)\).

\subsection{Brain Aging Trajectory Classification}\label{sec:class}
It is challenging to perform downstream analyses directly on INRs as they may comprise a large number of parameters $\theta_m$ (on the order of millions). Then, even simple operations such as applying an MLP to flattened parameters can lead to prohibitive memory requirements and result in poor generalization due to permutation and scale symmetries in parameter space \cite{hecht1990algebraic}.

Therefore, we propose to efficiently solve this brain aging trajectory classification challenge in two steps. First, we aim to reduce the effects of parameter symmetries by using a data-driven initialization strategy where we optimize an INR \(f_{\theta^*}(\cdot)\) across all subjects. The resulting initial parameters \(\theta^*\) are optimized for reconstructing an average dataset across the temporal range and can then be quickly adapted to form a subject-specific representation \(\theta_m\). Second, we implement a strategy to organize parameters $\theta_m$ for efficient processing and use an encoder to map them to a latent space optimized for classification. For each hidden layer of the INR network, we stack the weights and biases, \(\mathbf{W}_l\) and \(\mathbf{b}_l^{T}\), to produce layer-specific transformation parameters \(\mathbf{P}_{l}\in \mathbb{R}^{(H+1)\times H}\). These are then grouped by their respective processing streams to form spatial parameters \(\mathbf{P}_{\text{space}}\in \mathbb{R}^{L_{S}(H+1)\times H}\), temporal parameters \(\mathbf{P}_{\text{time}}\in \mathbb{R}^{L_{T}(H+1)\times H}\), and space-time combined parameters \(\mathbf{P}_{\text{com}}\in \mathbb{R}^{L_{C}(H+1)\times H}\), for \(L_{S}\), \(L_{T}\), and \(L_{C}\) number of layers in each stream. To build an efficient classifier, we employ an encoder inspired by \cite{luigi2023deep} consisting of three blocks of linear layers with batch norm and ReLU activation (see Fig. \ref{fig:wf}). This encoder processes the input parameters matrix row-wise, akin to processing each neuron's weight vector or each layer's bias vector independently, while sequentially expanding the dimension \(H\) to \(H'\). Therefore, the same encoder parameters are learned and used to process each row of INR parameters. A latent representation \(\mathbf{l}\in \mathbb{R}^{H'}\) is obtained by max pooling across each column, and is then passed on through fully connected layers, with ReLU activation and dropout, which are trained for trajectory classification using a binary cross-entropy loss. Ultimately, this process enables us to build a classifier that is capable of processing any combination of INR parameter streams.

\subsection{Brain Aging Trajectory Simulation}\label{sec:sim}
For an initial proof-of-concept evaluation, we design a new framework for simulating brain aging trajectories to remove the effects of uncontrollable imperfections (\textit{e.g.,} registration errors and misalignment, movement artifacts, scanner-related biases) associated with real longitudinal imaging data. Specifically, we develop a method for artificially generating a sequence of longitudinal 3D MRI data with accelerated, AD-like brain aging by systematically altering data from simulated subjects with healthy brain aging. This leads to perfect counterfactual scenarios where a simulated subject with the same base morphology can be rendered with either healthy or pathological aging, thus allowing us to identify morphology-driven classification bias.

Therefore, we first simulate subject-specific trajectories by sampling longitudinal MRI data from a recent 3D (age-)conditional diffusion model \cite{wilms2025lightweight} trained on cross-sectional data from more than 5,000 cognitively healthy subjects in the UK Biobank. This model has shown a competitive performance for biological brain age prediction and excels at generating high-resolution, realistic 3D T1-weighted MR images for given chronological age values. With this diffusion model, we are able to generate diverse longitudinal data for artificial subjects between 50 and 90 years of age with any temporal sampling scheme.

\begin{figure}[]
\includegraphics[width=\textwidth]{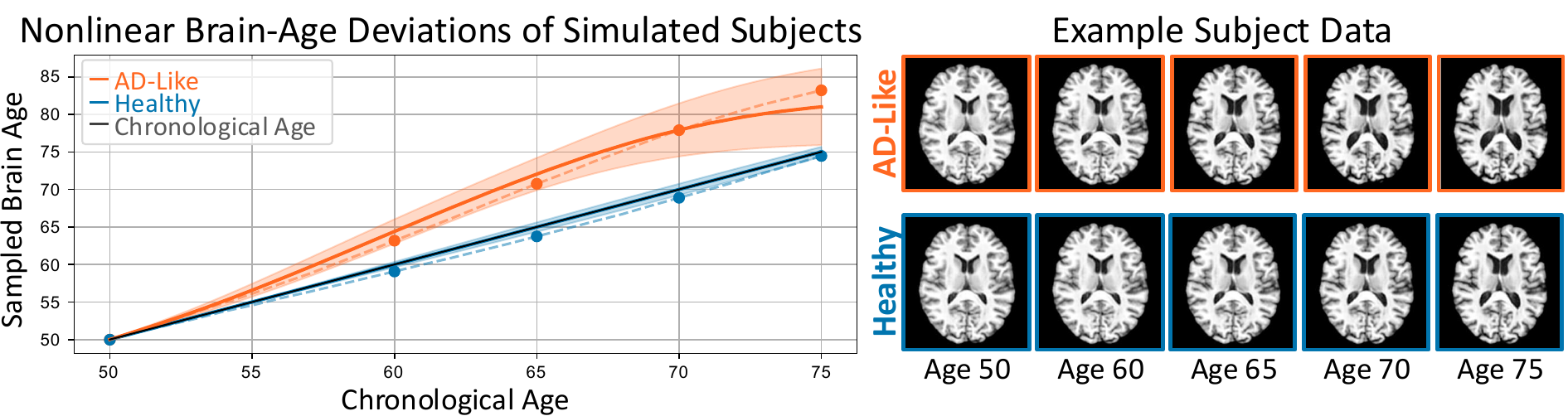}
\caption{Left: Exemplary simulated brain age deviation mappings (dashed lines) from our nonlinear ODE model. Solid lines denote the mean mapping and shaded regions indicate standard deviation across 450 simulations. Right: Visualization of sampled subject data as both healthy and AD-like trajectories.} \label{fig:traj}
\end{figure}

Given its training data, the diffusion model only generates healthy aging trajectories. To simulate a trajectory classification task (healthy \textit{vs.} AD-like accelerated aging), we systematically alter the sampled healthy trajectories by introducing a non-linear deviation mapping between chronological age and biological brain age. In the diffusion model-sampled data, both values are equal and we mimic the brain age drift commonly observed in AD \cite{franke2019ten} by reassigning biologically older brains to lower chronological ages. Driving this deviation in a realistic manner, we model the change in biological brain age, \(\frac{dBA}{dt}\), in relation to chronological age $t$ with an ordinary differential equation (ODE), such that:
\begin{equation}
    \frac{d\,BA}{dt} = 1 + \alpha \cdot \left(\frac{1}{1 + e^{-r\,(t - t_{\text{start}})}} - \frac{2}{1 + e^{-r\,(t - t_{\text{end}})}}\right) + \varepsilon(t),
\end{equation}
with initial condition $BA(t_0) = t_0$ (biological brain age equals chronological age). This ODE comprises two sigmoidal terms inducing nonlinear, age-dependent acceleration and deceleration of brain aging. Here, $\alpha$ controls the strength of brain aging, $r$ is the transition rate, and $\varepsilon$ introduces biological variability and measurement uncertainty \cite{franke2019ten}. To simulate AD-like aging trajectories, we introduce a gradual increase in brain age centred around \(t_{\text{start}}\) and then invoke dampening in later years, at \(t_{\text{end}} \,(\text{where }t_{\text{start}}<t_{\text{end}})\), to model saturation of disease progression and to prevent the sampled brain ages from exceeding those the diffusion model generates. To produce AD-like trajectories where brain age surpasses chronological age (\textit{e.g.,} by 5-10 years at age 80 \cite{gaser2013brainage,habes2021brain}), we randomly sample parameters as \( \alpha \sim \mathcal{N}(1.10, 0.05), \,r \sim \mathcal{N}(0.25, 0.05),\,\text{and }t_{\text{start}} \sim \mathcal{N}(55, 2.5)\). We also slightly alter the original healthy trajectories using this deviation model to simulate additional biological variability. Those parameters are sampled from \( \alpha \sim \mathcal{N}(0.00, 0.10), \,r \sim \mathcal{N}(0.25, 0.05),\,\text{and }t_{\text{start}} \sim \mathcal{N}(55, 1.0)\), which are consistent with literature indicating negligible systematic deviation from chronological age \cite{franke2019ten}. Exemplary deviation functions are visualized in Fig. \ref{fig:traj}.

\subsection{Experimental Setup}\label{sec:exp}
To evaluate the brain aging trajectories that we model, we build and classify INRs using \textit{regular} temporal sampling (with a fixed number of scans and same acquisition times across all subjects) and \textit{irregular} temporal sampling (with a different number of scans and acquisition times per subject). All experiments were conducted on a single Nvidia RTX4090 GPU with 24 GB. Code is available at \url{https://github.com/AgamAulakh/INR-Trajectory-Classification}.

\textbf{Data:} The diffusion model described in Sec. \ref{sec:sim} is used to generate annual scans \(\{I_{0},\dots,I_{N}\}\) of 450 simulated subjects at chronological ages \(\{t_{0},\dots,t_{N}\}\), where \(t_0=50\) and \(t_N=90\). This 3D MRI data is centre-cropped to dimensions of \(147\times183\times169\) voxels with 1 mm isotropic voxel spacing. The 450 simulated subjects are divided into train and test sets following a 2:1 ratio, and for all subjects, the ages associated with each sampled scan are reassigned using the ODE model described in Sec. \ref{sec:sim}. Importantly, we include both trajectory pairs (healthy \textit{and} AD-like) for subjects in the test set, but keep only one instance of a subject (healthy \textit{or} AD-like) in the training data. Thus, although we do not train any classifier using paired trajectory data, the inclusion of both healthy and AD-like subject counterparts in the test set allows us to assess how different aging trajectories are interpreted as they act on the same brain morphology. With this simulated data, we consider a \textit{regular} sampling scheme where, for all subjects, we select four scans at fixed chronological ages \(\{50, 58, 67, 75\}\) for trajectory classification. We also consider a second, more realistic, \textit{irregular} sampling scheme, where we randomly select three to five scans per subject within the range of 50-75 years.

\textbf{INR Fitting:} The INR architecture consists of eight residually connected layers of hidden size 512, with the spatial and temporal streams kept separate for the first five layers. The initialization INR \(\theta^*\) is pretrained for 250 iterations, where we sample 0.1\% of voxels from a batch size of three subjects per iteration. Then, for each subject, we finetune the initialization for 100 additional iterations to construct subject-specific INRs \(\theta_m\) by sampling 1\% of voxels. This process is followed for both sampling experiments independently. Initialization training takes less than one hour, and finetuning requires about 30 seconds per subject.


\textbf{Classification:} We train the combined encoder-classifier introduced in Sec. \ref{sec:class} for the regular and irregular sampling INRs independently. Processing the INR parameters row-wise, the encoder sequentially expands the dimensionality through linear-batch norm-ReLU blocks of size 512, 1024, and 2048. The resulting latent representation after max pooling is then passed through two fully connected layers for classification. This model is trained for 100 epochs. To assess the advantages of our semi-disentangled INR architecture, we evaluate trajectory classification performance across single-stream (\textit{e.g.,} $\mathbf{P}_{\text{space}}$) and multi-stream INR parameters (\textit{e.g.,} $\mathbf{P}_{\text{time}}+\mathbf{P}_{\text{com}}$).

For initial baseline comparisons, we select a variant of the Simple Fully Convolutional Network (SFCN), which has been shown to achieve state-of-the-art performance for brain age prediction \cite{peng2021accurate} and similarly strong results within the context of Alzheimer's disease classification \cite{early2025comparison}. We adapt the SFCN architecture used in \cite{early2025comparison} to process multiple scans simultaneously as multi-channel inputs. For the regular sampling experiment, the SFCN receives all four scans stacked in chronological order in the channel dimension. In the irregular sampling experiment, the SFCN receives the scans in chronological order with zero-padding applied if fewer than five scans are available.

\section{Results and Discussion}
We first assess the reconstructions produced by our semi-disentangled INR architecture when fit to \textit{irregular} temporal data, and then present trajectory classification results across both temporal sampling experiments. 

\begin{figure}[]
\includegraphics[width=\textwidth]{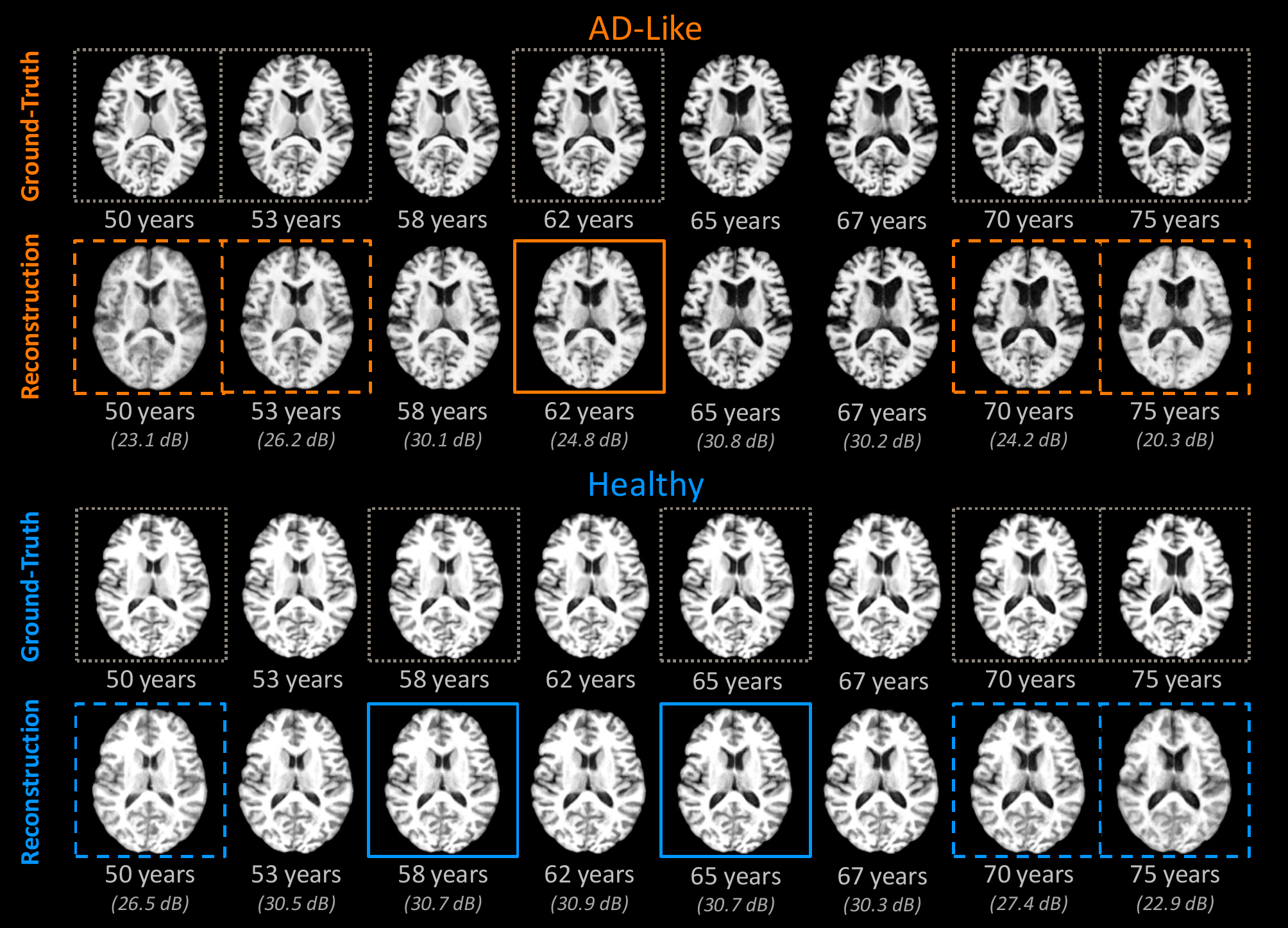}
\caption{Examples of reconstructed scans for simulated subjects with AD-like (orange) and healthy (blue) brain aging trajectories. For both AD-like and healthy trajectories, the top row shows the ground-truth scans and the bottom row shows their corresponding reconstructions. Grey dotted outlines indicate data not used for training. Solid outlines (orange/blue) indicate interpolated reconstructions and dashed outlines (orange/blue) indicate extrapolated reconstructions.} \label{fig:reco}
\end{figure}

\textbf{Quality of INR Reconstructions:} Using our architecture (Sec. \ref{sec:exp}), we produce INRs with a maximum of 4.20 million parameters (\(\mathbf{P}_{space}+\mathbf{P}_{time}+\mathbf{P}_{com}\)), which results in data compression ranging from 18-30\% for datasets with five to three scans each. Using subject-specific INRs trained with \textit{irregular} temporal sampling, we reconstruct imaging data across all yearly timepoints included within our simulation. Tab. \ref{tab:reco} summarizes the reconstruction data quality across timepoints used for INR finetuning (training data) and those not included (interpolated/extrapolated timepoints). We observe lower quality for reconstructions generated through extrapolations, which may indicate that the INRs default to producing the average trajectory learned during initialization for timepoints where prior or subsequent subject data is not available. Reconstructions from INRs of two subjects are shown in Fig. \ref{fig:reco}, demonstrating that the INRs can maintain subject-specific morphology and produce a stable trajectory across the temporal range without closely overfitting to (\textit{i.e.,} only producing viable reconstructions of) timepoints used for finetuning.

\begin{table}
\caption{Quality of reconstructed training data, interpolations, and extrapolations from INRs in the \textit{irregular} sampling experiment. We report Mean Square Error (MSE), Peak Signal-to-Noise Ratio (PSNR), and Structural Similarity Index Measure (SSIM).}\label{tab:reco}
\centering
\begin{tabular}{lccccccc}\toprule
\multicolumn{1}{c}{}  &\multicolumn{3}{c}{\textbf{Healthy Subjects}}  && \multicolumn{3}{c}{\textbf{AD-Like Subjects}}  \\
\cmidrule{2-4} \cmidrule{6-8} 
\textbf{Reconstruction}&MSE& PSNR& SSIM& &MSE& PSNR& SSIM\\
\midrule
 Training Data& 3.95 \(\times10^{-3}\) & 30.2 & 0.953 &  & 3.91 \(\times10^{-3}\) & 30.2 & 0.952\\
 Interpolation& 5.00 \(\times10^{-3}\) & 29.3 &0.943 &  & 6.63 \(\times10^{-3}\) & 28.3 &0.930\\
 Extrapolation& 10.6 \(\times10^{-3}\) & 26.1 &0.898 &  & 12.1 \(\times10^{-3}\) & 25.6 &0.887\\\midrule
\end{tabular}
\end{table}

\textbf{Brain Aging Trajectory Classification:} Tab. \ref{tab:class} shows classification results across \textit{regular} and \textit{irregular} sampling experiments. We achieve consistently strong classification performance using $\mathbf{P}_{\text{time}}$ and $\mathbf{P}_{\text{time}}+\mathbf{P}_{\text{com}}$ INR parameters. In particular, these two parameter sets outperform the SFCN baseline in the irregular sampling scheme, which highlights the advantages of encoding longitudinal data as continuous representations. Furthermore, we observe chance-level classification accuracies using $\mathbf{P}_{\text{space}}$ across both sampling schemes, and show that excluding these parameters leads to notable improvements (\textit{e.g.,} 77\% when using \(\mathbf{P}_{\text{space}}+\mathbf{P}_{\text{time}}+\mathbf{P}_{\text{com}}\) \textit{vs.} 81\% with \(\mathbf{P}_{\text{time}}\)). This highlights the benefits of our semi-disentangled INR architecture, as we can selectively exclude parameters which may introduce spurious correlations that do not generalize beyond training data and, as a result, construct parameter-efficient representations. While we observe the highest accuracies using $\mathbf{P}_{\text{time}}$, different parameter combinations may be useful for other scenarios where, for example, one is interested in studying brain dynamics with acute and spatially-localized temporal changes.

\begin{table}[]
\caption{Classification results across single- and multi-stream INR parameters with SFCN benchmark. Number of parameters used for classification is reported in the order of millions (M). $\mathbf{P}_{\text{space}}$, $\mathbf{P}_{\text{time}}$, and $\mathbf{P}_{\text{com}}$ are abbreviated as $\mathbf{P}_{\text{s}}$, $\mathbf{P}_{\text{t}}$, and $\mathbf{P}_{\text{c}}$, respectively.}\label{tab:class}
\centering
\begin{tabular}{lccccc}
\toprule
\multicolumn{1}{c}{} &\multicolumn{2}{c}{\textbf{Regular Sampling}}  && \multicolumn{2}{c}{\textbf{Irregular Sampling}}  \\
\cmidrule{2-3} \cmidrule{5-6}
\textbf{Model}&Input Size (\(\downarrow\))& Accuracy (\(\uparrow\))&& Input Size (\(\downarrow\))& Accuracy (\(\uparrow\))\\
\midrule
SFCN & 18.1M & \textbf{97.7} && 22.7M & 73.7 \\
INR \((\mathbf{P}_{s})\)& 2.10M & 50.0 && 2.10M & 49.8\\
INR \((\mathbf{P}_{t})\)& \textbf{1.05M} & 95.6 && \textbf{1.05M} & \textbf{81.3}\\
INR \((\mathbf{P}_{s}+\mathbf{P}_{c})\)& 3.15M & 91.7 && 3.15M & 74.6 \\
INR \((\mathbf{P}_{t}+\mathbf{P}_{c})\)& 2.10M & 96.3 && 2.10M & 78.3 \\
INR \((\mathbf{P}_{s}+\mathbf{P}_{t}+\mathbf{P}_{c})\)& 4.20M & 95.3 && 4.20M & 77.3 \\
\bottomrule
\end{tabular}
\end{table}
To keep comparisons consistent, we do not provide the INR encoder and the SFCN with temporal information associated with the imaging data. However, it is implicitly encoded in the INR parameters during finetuning. This may be one reason why the SFCN performs poorly with irregular sampling, and, therefore, our use of this model may not provide the most competitive benchmark in this initial evaluation. Despite this, our INR-based results are highly promising and motivate immediate future work to use real data with more advanced baselines.

\section{Conclusion}
This paper presents a novel method for modelling and, for the first time, classifying brain aging trajectories using spatiotemporal INRs. We introduce a new INR architecture capable of partially disentangling spatiotemporal parameters and highlight its advantages in isolating informative features for the classification of healthy and AD-like brain aging trajectories. Using a newly designed trajectory simulation framework, we demonstrate that our proposed INR architecture can represent high-resolution 3D longitudinal MRI data with both regular and irregular temporal sampling, capture structural brain aging dynamics, and be directly applied in a downstream classification task. A notable strength of our approach is the flexibility it offers, both in adapting to data with varying sampling patterns and in its capacity to operate directly on neuroimaging data to disentangle spatial and temporal characteristics. Extending the analysis provided in this paper, future work is required to test this approach on real data.

\begin{credits}
\subsubsection{\ackname} This work was supported by the Department of Pediatrics and the Azrieli Accelerator at the University of Calgary, the Hotchkiss Brain Institute, the Alberta Children's Hospital Foundation, and the Natural Sciences and Engineering Research Council of Canada (RGPIN-03532-2023). 
\subsubsection{\discintname} The authors have no competing interests to declare.

\end{credits}

\bibliographystyle{splncs04}
\bibliography{Paper-0010}

\end{document}